\documentclass[11pt]{article}

\usepackage[preprint]{acl}

\usepackage{times}
\usepackage{latexsym}

\usepackage[T1]{fontenc}

\usepackage[utf8]{inputenc}

\usepackage{microtype}

\usepackage{inconsolata}

\usepackage{graphicx}

\usepackage{xspace}
\usepackage{amsmath}
\usepackage{booktabs}
\usepackage{tabularx}
\usepackage{multirow}

\usepackage{listings}
\usepackage{float}
\usepackage{algorithm}
\usepackage{algpseudocode}
\usepackage[table]{xcolor}

\usepackage{tcolorbox}        
\usepackage{amsmath}          
\usepackage{enumitem}         
\usepackage{xcolor}           
\usepackage{listings}         
\usepackage{float}            

\usepackage{xspace}
\usepackage{subcaption}
\tcbuselibrary{listings}
\usepackage{listings}
\usepackage{multicol} 
\usepackage{footnote} 
\usepackage[dvipsnames]{xcolor}
\usepackage{enumitem}
\usepackage{setspace}
\usepackage{soul}
\usepackage{wasysym} 
\usepackage{pifont} 

\usepackage{tcolorbox}
\tcbuselibrary{skins,breakable}
\usepackage{cuted}
\usepackage{xurl}
\usepackage{relsize}

\definecolor{DeltaBg}{HTML}{D4F2D7} 
\definecolor{SearchBg}{HTML}{C2E6F5} 
\definecolor{AgenticBg}{HTML}{F5C2CC} 
\definecolor{MathBg}{HTML}{E6D4F2} 
\definecolor{ScienceBg}{HTML}{FBE0BC} 
\definecolor{warning}{RGB}{210, 40, 40}
\definecolor{darkgreen}{RGB}{0, 181, 18}
\definecolor{darkred}{RGB}{252, 90, 90}

\newcommand{\mygreen}[1]{\cellcolor{darkgreen!#1}}
\newcommand{\gain}[1]{$\uparrow$ #1}

\definecolor{toolcardbox}{RGB}{240, 248, 255} 
\definecolor{toolcardborder}{RGB}{52, 52, 173} 

\newtcolorbox{promptbox}[2][]{
    colback=toolcardbox,
    colframe=toolcardborder,          
    coltitle=white,                
    arc=1pt,                       
    boxrule=1pt,
    fonttitle=\bfseries,
    title=#1,                      
    left=5pt,
    right=5pt,
    top=5pt,
    bottom=5pt,
    before skip=1em,
    after skip=1em,
    fontupper=\small,               
    breakable,     
    width=1.\linewidth, 
    #1                        
}

\newtcblisting{codebox}{
    colback=toolcardbox,
    colframe=white,
    arc=0pt,
    boxrule=0pt,
    listing only,
    listing options={
        language=Python,
        basicstyle=\ttfamily\footnotesize,
        keywordstyle=\color{blue!70!black}\bfseries,
        commentstyle=\itshape\color{green!40!black},
        stringstyle=\color{red!60!black},
        showstringspaces=false,
        columns=flexible,
        breaklines=true,
        inputencoding=utf8,
        extendedchars=true,
        texcl=true,
        escapeinside={(*@}{@*)},  
    },
    left=0pt,
    right=0pt,
    top=-10pt,
    bottom=-10pt,
    breakable,
    fontupper=\small,
}

\newcommand{\smalltt}[1]{{\ttfamily\fontsize{8}{10}\selectfont #1}}

\newcommand{\dataset}{\textsc{EvolveCoder-22k}\xspace}
\newcommand{\coder}{\textsc{EvolveCoder-4B}\xspace}

\title{EvolveCoder: Evolving Test Cases via Adversarial Verification for Code Reinforcement Learning}

\author{
 \textbf{Chi Ruan\textsuperscript{1}},
 \textbf{Dongfu Jiang\textsuperscript{1,2}},
 \textbf{Huaye Zeng\textsuperscript{3}},
 \textbf{Ping Nie\textsuperscript{1}},
 \textbf{Wenhu Chen\textsuperscript{1,2}},
\\
\\
 \textsuperscript{1}University of Waterloo,
 \textsuperscript{2}Vector Institute,
 \textsuperscript{3}Harvard University,
\\
{\relsize{-0.5}\texttt{cruan059@uottawa.ca}}
\vspace{5pt}
 \\
 \vspace{0pt} 
  \url{https://github.com/TIGER-AI-Lab/EvolveCoder}
}

\begin{document}
\maketitle
\begin{abstract}


Reinforcement learning with verifiable rewards (RLVR) is a promising approach for improving code generation in large language models, but its effectiveness is limited by weak and static verification signals in existing coding RL datasets. In this paper, we propose a solution-conditioned and adversarial verification framework that iteratively refines test cases based on the execution behaviors of candidate solutions, with the goal of increasing difficulty, improving discriminative power, and reducing redundancy. Based on this framework, we introduce \dataset{}, a large-scale coding reinforcement learning dataset constructed through multiple rounds of adversarial test case evolution. Empirical analysis shows that iterative refinement substantially strengthens verification, with pass@1 decreasing from $43.80$ to $31.22$. Reinforcement learning on \dataset{} yields stable optimization and consistent performance gains, improving Qwen3-4B by an average of $4.2$ points across four downstream benchmarks and outperforming strong 4B-scale baselines. Our results highlight the importance of adversarial, solution-conditioned verification for effective and scalable reinforcement learning in code generation.

\end{abstract}

\section{Introduction}
Large language models (LLMs) have recently demonstrated remarkable advances in coding and logical reasoning, exemplified by systems such as OpenAI’s o1–o4~\citep{jaech2024openai}, DeepSeek-R1~\citep{guo2025deepseek}, and Kimi-K2~\citep{kimik2}. These models achieve strong performance across a wide range of challenging programming benchmarks. A key driver behind this progress is the growing adoption of reinforcement learning with verifiable rewards (RLVR) as a post-training paradigm, where rewards are defined by objective and externally verifiable criteria. When combined with chain-of-thought (CoT) reasoning~\citep{wei2022chain}, RLVR encourages models to develop more reliable intermediate reasoning processes rather than merely optimizing final outputs. Code generation is particularly well suited to RLVR, as candidate solutions can be precisely verified through automated test cases, enabling scalable and reliable reward signals. As a result, RLVR has become a central mechanism in modern coding-oriented LLMs, underpinning a series of recent state-of-the-art systems~\citep{katcoder, ruan2025critique, roziere2023code, guo2024deepseek}.

Despite the effectiveness of RLVR in recent coding LLMs, its success ultimately hinges on the quality of verification signals—specifically, the quality of test cases. However, existing coding RL datasets remain poorly aligned with the requirements of reliable and informative reward supervision. 
\textbf{(1)} Human-annotated datasets such as TACO~\citep{li2023taco}, APPS~\citep{hendrycksapps2021}, and CodeContests~\citep{codecontest} frequently \emph{fail to expose critical corner cases}~\citep{tong2024codejudge}, resulting in weakly discriminative and incomplete reward signals. 
\textbf{(2)} Recent approaches, including HardTest~\citep{he2025hardtests} and rStarCoder~\citep{liu2025rstar}, attempt to augment test suites by automatically generating additional inputs and executing reference solutions to obtain corresponding outputs. However, these methods often \emph{lack principled filtering}, leading to high verification cost and limited scalability~\citep{ruan2025critique}. 
\textbf{(3)} By contrast, single-pass generative pipelines such as AceCoder~\citep{zeng2025acecoder} construct problems and test cases without explicit incentives to target model failure modes, causing \emph{corner cases to emerge only sporadically} rather than being systematically induced. 
As a result, existing coding RL datasets struggle to provide verification signals that are simultaneously reliable, adversarial, and computationally tractable, ultimately constraining the effectiveness of RL for code generation.

Moreover, RLVR has been observed to suffer from \emph{vanishing advantage} issues~\citep{Yu2025DAPOAO,Su2025PixelRI} due to imbalanced task difficulty in existing coding RL datasets. When problems are either too easy or too difficult, policy models fail to generate solutions that meaningfully differ in reward outcomes, leading to weak or indistinguishable learning signals. This issue becomes increasingly severe as reinforcement learning proceeds over more optimization steps, further reducing training efficiency and stability.

Motivated by these limitations of static and single-pass verification, we argue that effective test case construction should be \textbf{adversarially conditioned on the execution outcomes of candidate solutions}, rather than generated solely from the problem statement or a reference solution. Our key insight is that candidate solutions and their execution behaviors directly reveal which aspects of program semantics remain insufficiently challenged or indistinguishable under the current test suite. Leveraging this insight, we propose a solution-conditioned verification paradigm that adversarially refines test cases based on observed model behaviors, with the dual goals of \emph{increasing test difficulty} and \emph{improving discriminative power}. To ensure scalability and robustness, we further discourage redundant verification that repeatedly targets the same failure patterns, encouraging broader coverage of distinct vulnerabilities. Together, these principles enable the construction of verification signals that are more informative, reliable, and effective for reinforcement learning.

Based on this paradigm, we introduce \dataset{}, a new coding reinforcement learning dataset that augments existing programming problems with substantially stronger verification signals through multiple rounds of adversarial evolution. \dataset{} pairs program instances with harder and more discriminative test suites, generated by conditioning on candidate solutions and their execution outcomes rather than relying on static, single-pass procedures. Compared to prior coding RL datasets, the resulting verification suites exhibit reduced redundancy and broader coverage of distinct failure patterns, yielding more informative execution-based rewards. Consequently, \dataset{} is well suited for stable and effective reinforcement learning for code generation while remaining computationally tractable at scale.

We conduct detailed dataset analysis to study how test cases and program pass rates evolve over four rounds of verification, with pass@1 being reduced from 43.80 to 31.22 due to stronger verification, highlighting the increasing difficulty induced by iterative adversarial refinement. We further train \coder{} on \dataset{} and observe an average improvement of $4.2$ points across four downstream coding benchmarks compared to the Qwen3-4B starting point, and a $1.8$-point gain over the strongest 4B baseline, Critique-Coder~\citep{ruan2025critique}. Comprehensive ablation studies demonstrate that increasing the number of verification evolution rounds yields more stable optimization and consistently improves code generation performance. We hope our work highlights the importance of solution-conditioned and adversarial verification for advancing reinforcement learning in code generation.

\section{Dataset Construction}

\begin{figure*}[!t]
  \centering
  \includegraphics[width=0.8\linewidth]{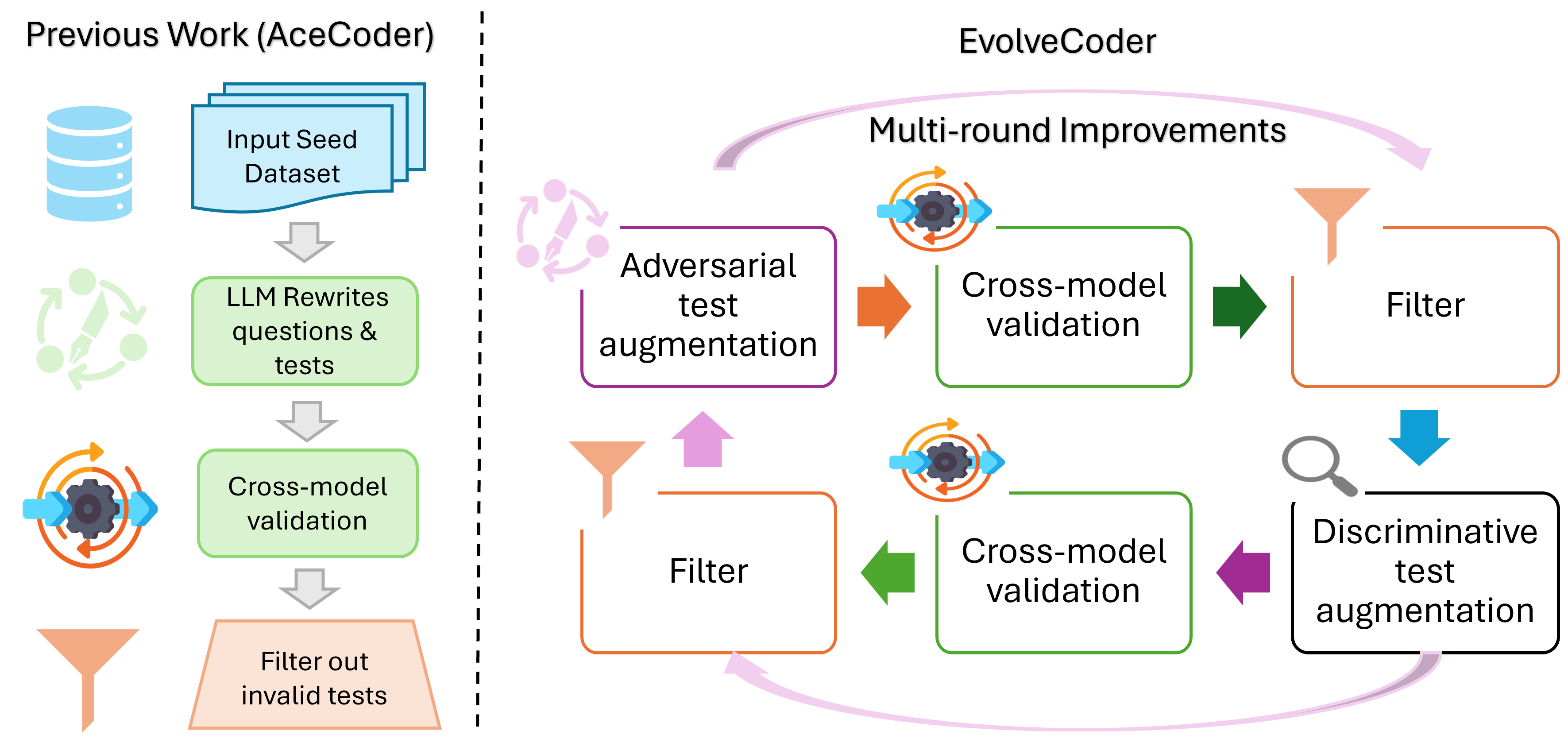}
  \caption{\dataset Construction Pipeline}
  \label{fig:main_graph}
  \vspace{0em}
\end{figure*}

We construct our dataset through a multi-stage pipeline that progressively strengthens problem specifications and verification signals, as illustrated in Figure~\ref{fig:main_graph}. Starting from curated seed problems, we refine task formulations and iteratively generate and filter test cases by leveraging execution feedback from diverse candidate solutions. This process produces fine-grained and reliable verification signals suitable for reinforcement learning with verifiable rewards.

\begin{table}[b]
\centering
\caption{Statistics of seed datasets before and after filtering, and the final dataset \dataset.
Filtering removes highly similar problem instances based on embedding similarity.}
\resizebox{\linewidth}{!}{
\begin{tabular}{lccc}
\toprule
\textbf{Dataset} 
& \textbf{Before filtering} 
& \textbf{After filtering} 
& \textbf{\dataset} \\
\midrule
TACO            & 26,433 & 13,776 & 9,099 \\
APPS            & 8,765  & 2,295  & 1,925 \\
PrimeIntellect  & 16,252 & 5,021  & 3,702 \\
Contests        & 13,610 & 4,386  & 3,060 \\
Codeforces      & 10,024 & 4,959  & 3,856 \\
\midrule
\textbf{Overall} & \textbf{75,084} & \textbf{30,437} & \textbf{21,642} \\
\bottomrule
\end{tabular}}
\label{tab:filtering_stats}
\end{table}

\subsection{Seed Datasets Construction}

To build a large-scale and curated coding dataset, we start by collecting high-quality seed datasets. We aggregate seed data from five publicly available datasets, including TACO \citep{li2023taco}, APPS \citep{hendrycksapps2021}, SYNTHETIC-1 \citep{2025synthetic1}, Codeforces \citep{penedo2025codeforces}, and CodeContests \citep{codecontest}. These datasets provide natural language problem descriptions paired with reference implementations in Python, and cover a wide range of domains and difficulty levels. The resulting collection serves as the foundation for all subsequent stages of our pipeline.

\begin{figure}[t]
  \centering
  \includegraphics[width=0.9\linewidth]{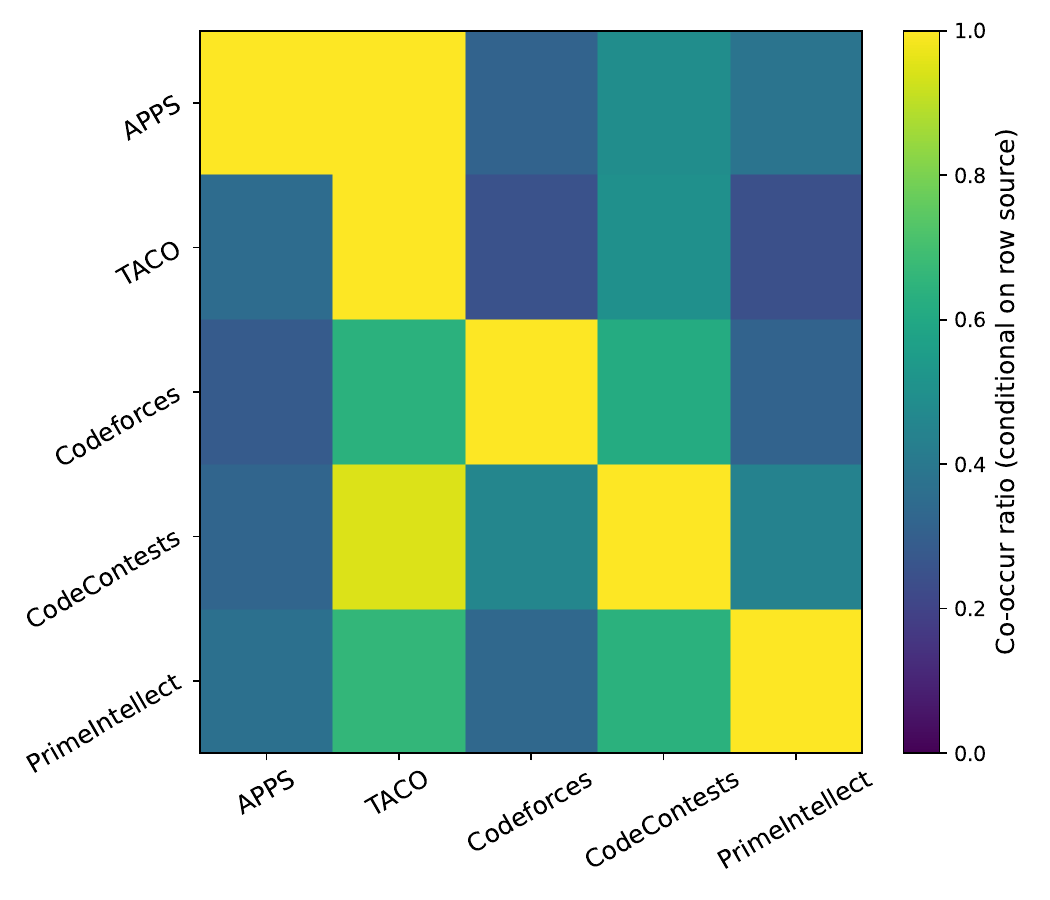}
  \caption{Semantic duplication across dataset sources.}
  \label{fig:cross_source_dup}
  \vspace{0em}
\end{figure}

\paragraph{Problems De-duplication}
Aggregating problems from multiple sources inevitably introduces redundancy due to shared provenance across competitive programming platforms and educational repositories. Figure~\ref{fig:cross_source_dup} presents cross-source semantic duplication in terms of relative overlap ratios between datasets, with diagonal entries normalized to one by definition. The most pronounced cross-dataset similarity is observed between APPS and TACO, as well as between TACO and CodeContests, indicating substantial reuse of semantically equivalent or closely related problems across these sources. These patterns suggest that redundancy primarily arises from problem redistribution across benchmarks rather than isolated dataset construction. If left unaddressed, such cross-source duplication can distort the effective data distribution and reduce the true diversity of training signals.

\begin{figure}[t]
  \centering
  \includegraphics[width=\linewidth]{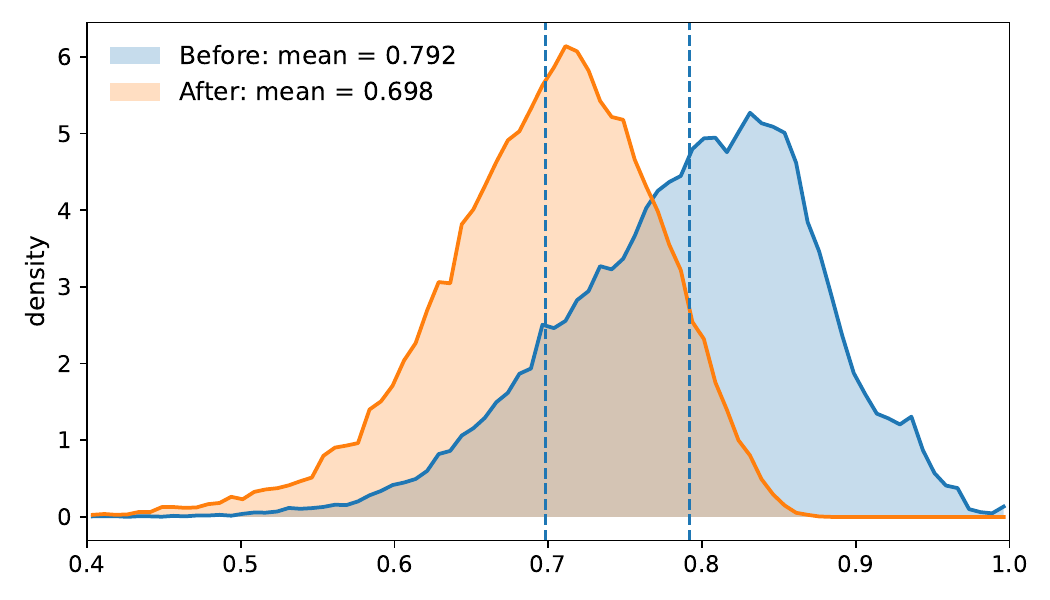}
  \caption{Distribution of mean cosine similarity to the 10 nearest neighbors before and after filtering.}
  \label{fig:cosine_dist}
  \vspace{0em}
\end{figure}

To address this issue, we perform semantic-level deduplication based on dense sentence embeddings. We encode all problem descriptions using the pretrained \textsc{all-mpnet-base-v2} model~\citep{allmpnetbasev2} and compute pairwise cosine similarity between embeddings. For pairs with similarity greater than 0.9, we randomly retain a single representative problem and discard the rest. Table~\ref{tab:filtering_stats} summarizes the dataset statistics before and after filtering, showing that a substantial fraction of redundant problems is removed across all sources, with retention rates ranging from 30.89\% in PrimeIntellect to 52.12\% in TACO. To illustrate the effect of filtering, Figure~\ref{fig:cosine_dist} shows the distribution of mean cosine similarity to each problem’s 10 nearest neighbors before and after filtering. After filtering, the distribution shifts markedly toward lower similarity values, with the average similarity decreasing from 0.792 to 0.698. This shift indicates that local semantic neighborhoods become substantially less redundant after filtering. Together, these results show that our filtering procedure improves dataset diversity while preserving large-scale coverage, yielding a cleaner seed set for downstream data generation and reinforcement learning.

\subsection{Seed Dataset Refinement}
While seed datasets provide programming problems paired with verifiable test cases, many of these problems have become insufficiently challenging for modern reasoning-capable models. Moreover, test cases are typically represented as raw stdin--stdout pairs, where multiple input--output examples are concatenated into a single evaluation instance. This format obscures fine-grained execution outcomes and makes it difficult to assess how a solution behaves on individual test cases. Under these conditions, approaches \citep{he2025hardtests, liu2025rstar} that aim to induce harder corner cases often rely on large-scale randomized input generation, with outputs obtained by executing oracle solutions, followed by limited or post-hoc filtering. In practice, this process still results in substantial redundancy, as many test cases probe the same underlying failure modes. In practice, such redundancy significantly increases evaluation cost and training time during reinforcement learning.

To address these limitations, we adopt the seed refinement strategy introduced in AceCoder~\citep{zeng2025acecoder} as an initial preprocessing step. Given a filtered question--solution pair $(p, c)$, where $p$ denotes an original programming problem and $c$ a reference code solution, we prompt \textsc{GPT-4.1mini}~\citep{openaio1} to generate a refined LeetCode-style problem $q$ with a rewritten problem specification and approximately 20 independently executable test cases $\{t_1, \dots, t_m\}$. These test cases are explicitly structured and self-contained, enabling fine-grained evaluation of candidate solutions across diverse behavioral regimes. By increasing problem difficulty while reducing redundancy among test cases, this refinement step yields a cleaner and more discriminative seed dataset, providing a stronger foundation for downstream test generation and reinforcement learning. See Appendix~\ref{appendix:seed_code_prompt} for the refinement prompt.

\subsection{Progressive Test Case Refinement} \label{Progressive Test Case Refinement}
\textbf{Pipeline Overview.} We propose a progressive test case refinement pipeline that iteratively strengthens verification signals by conditioning test generation on diverse candidate solutions and their execution outcomes. Starting from an initial test suite, the pipeline alternates between adversarial test generation and discriminative test generation, progressively exposing unresolved failure modes while controlling redundancy and evaluation cost. After completing all refinement rounds, a final test suite filtering stage is applied to consolidate generated tests and produce a compact, reliable verification set.

\textbf{Diverse Solution Sampling.}
To capture a wide range of executable behaviors and failure patterns, we construct a behaviorally diverse pool of candidate solutions. Specifically, we use 8 independently trained open-source reasoning models, including the Qwen3 family (4B, 8B, 14B, 30B-A3B, and 32B) \citep{yang2025qwen3}, MiMo-7B-RL \citep{xia2025mimo}, Phi-4-Reasoning \citep{abdin2025phi4reasoning}, and DeepSeek-R1-Distill-Qwen-32B \citep{guo2025deepseek}. For each problem $q$, each model generates 8 solutions, yielding a total of 64 candidates denoted as $\mathcal{S}(q) = \{ s_i \}_{i=1}^{64}$. This multi-model sampling exposes diverse reasoning trajectories and systematic failure behaviors that are difficult to obtain from single-model generation and serves as the basis for subsequent test case refinement.

\textbf{Adversarial Test Case Generation.}
Building on the behaviorally diverse solution pool, we refine verification signals through adversarial test generation conditioned on candidate programs and their execution behaviors. Given a solution set $\mathcal{S}(q)$ and an initial test suite $\mathcal{T}(q)$, we evaluate all solutions on all tests to construct a binary pass matrix $\mathbf{M}$. Tests that are passed by nearly all solutions are removed, yielding a reduced suite $\widetilde{\mathcal{T}}(q)$ that better captures unresolved behavioral differences.

Based on evaluation outcomes on $\widetilde{\mathcal{T}}(q)$, we form a representative solution subset $\mathcal{S}^\star(q)$ by selecting the two highest-pass-rate solutions together with three solutions exhibiting maximal pairwise disagreement in pass–fail patterns, measured by Hamming distance. This selection balances solution quality and behavioral diversity. Conditioned on the problem description, $\widetilde{\mathcal{T}}(q)$, $\mathcal{S}^\star(q)$, and their fine-grained execution results, the test generation model synthesizes new assert-based tests $\mathcal{T}^{+}(q)$ that adversarially target uncovered corner cases, producing more informative and discriminative verification signals. See Appendix~\ref{appendix:adversarial_prompt} for the refinement prompt.







\begin{table*}[t]
\centering
\footnotesize
\setlength{\tabcolsep}{0.5em}
\renewcommand{\arraystretch}{0.95}
\caption{Dataset statistics across iterative construction rounds. For each round, we report the number of problems and the average number of test cases per problem for each subset.}
\resizebox{\textwidth}{!}{%
\begin{tabular}{@{}cc|cccccc@{}}
\toprule
\textbf{Round} & 
  \textbf{Metric} & 
  \textbf{Overall} & 
  \textbf{TACO} & 
  \textbf{APPS} & 
  \textbf{PrimeIntellect} & 
  \textbf{CodeContests} & 
  \textbf{Codeforces} \\
\midrule
\multirow{2}{*}{0}
& \#Problems   & 18{,}803 & 7{,}919 & 1{,}691 & 3{,}281 & 2{,}679 & 3{,}233 \\
& \#Avg. Tests & 10.76 & 10.70 & 10.78 & 10.49 & 10.77 & 11.20 \\
\midrule
\multirow{2}{*}{1}
& \#Problems   & 20{,}867 & 8{,}768 & 1{,}850 & 3{,}572 & 2{,}960 & 3{,}717 \\
& \#Avg. Tests & 20.87 & 20.43 & 20.13 & 19.45 & 21.17 & 23.39 \\
\midrule
\multirow{2}{*}{2}
& \#Problems   & 21{,}412 & 8{,}981 & 1{,}898 & 3{,}676 & 3{,}032 & 3{,}825 \\
& \#Avg. Tests & 28.39 & 27.58 & 26.70 & 25.84 & 29.16 & 32.96 \\
\midrule
\multirow{2}{*}{3}
& \#Problems   & 21{,}642 & 9{,}099 & 1{,}925 & 3{,}702 & 3{,}060 & 3{,}856 \\
& \#Avg. Tests & 35.04 & 33.80 & 32.17 & 31.43 & 36.38 & 41.81 \\
\bottomrule
\end{tabular}%
}
\label{tab:dataset_stats}
\end{table*}

\textbf{Discriminative Test Case Generation.}
Complementary to adversarial refinement, we introduce a discriminative test generation strategy targeting candidate programs that remain difficult to distinguish under existing verification signals. Based on the solution–test pass matrix $\mathbf{M}$, we first filter the test suite in two stages. We remove tests with extremely low pass rates across the solution pool, below 0.1, as such tests are likely erroneous and yield unreliable discrimination. We then discard tests with identical pass vectors, retaining a single representative per equivalence class. This produces a compact filtered suite $\widehat{\mathcal{T}}(q)$ that preserves the essential discriminative structure while reducing redundancy.

Using evaluation vectors on $\widehat{\mathcal{T}}(q)$, we select a subset of five behaviorally overlapping solutions $\mathcal{S}^{\sim}(q)\subset\mathcal{S}(q)$ by prioritizing solutions with identical or near-identical pass–fail patterns, and otherwise minimizing pairwise Hamming distance. Conditioned on the problem description, $\widehat{\mathcal{T}}(q)$, $\mathcal{S}^{\sim}(q)$, and their evaluation outcomes, the generator synthesizes new assert-based tests $\mathcal{T}^{\ddagger}(q)$ under an explicit split constraint, requiring each test to be passed by at least one solution and failed by at least one solution. This procedure exposes fine-grained behavioral differences among solutions that were previously indistinguishable. See Appendix~\ref{appendix:discriminative_prompt} for the refinement prompt.

\textbf{Test Suite Filtering.}
This stage refines the test suite $\mathcal{T}(q)$ accumulated over multiple refinement rounds by removing test cases with limited discriminative value across the candidate solution pool $\mathcal{S}(q)$. For each problem $q$, we analyze the test--solution pass matrix $\mathbf{M}$ defined over $\mathcal{T}(q)$ and $\mathcal{S}(q)$. We first remove tests with extremely low empirical pass rates, below 10\%, as such cases are likely to be unreliable. Next, tests with identical pass--fail vectors in $\mathbf{M}$ are grouped, and up to five representatives are retained per group to control redundancy. The resulting reduced test suite $\mathcal{T}'(q)$ is further constrained by discarding problems with fewer than five retained tests or with more than 60 perfectly solved solutions. This filtering procedure yields a smaller but more informative test suite that preserves meaningful behavioral differences while enabling efficient and reliable evaluation across refinement rounds.

\section{Dataset Analysis}

This section provides an empirical analysis of \dataset, the dataset obtained after multi-round test case refinement. We analyze its evolution across refinement rounds, focusing on dataset scale, test case retention, and difficulty progression, and evaluate verification quality and model performance on the final dataset.

\subsection{Dataset Scale Across Iterations}
Table \ref{tab:dataset_stats} summarizes the evolution of dataset scale across iterative construction rounds. We observe a monotonic increase in both the number of problems and the average number of test cases per problem. Overall, the dataset expands from 18,803 problems in Round~0 to 21,642 problems in Round 3, while the average number of test cases per problem increases from 10.76 to 35.04, reflecting a substantial growth in verification volume over successive iterations.

This growth pattern is consistent across all source subsets, including TACO, APPS, PrimeIntellect, CodeContests, and Codeforces. The reported problem counts correspond to the dataset state after the filtering step described in Section~\ref{Progressive Test Case Refinement}, which discards problems with fewer than five retained test cases or with an excessive number of perfectly solved solutions. The continued increase in problem count, therefore, indicates that newly generated problems and refined test suites consistently satisfy these quality constraints, allowing the dataset to scale in a balanced manner without concentrating expansion on any specific source.

\begin{figure}[t]
  \centering
  \includegraphics[width=\linewidth]{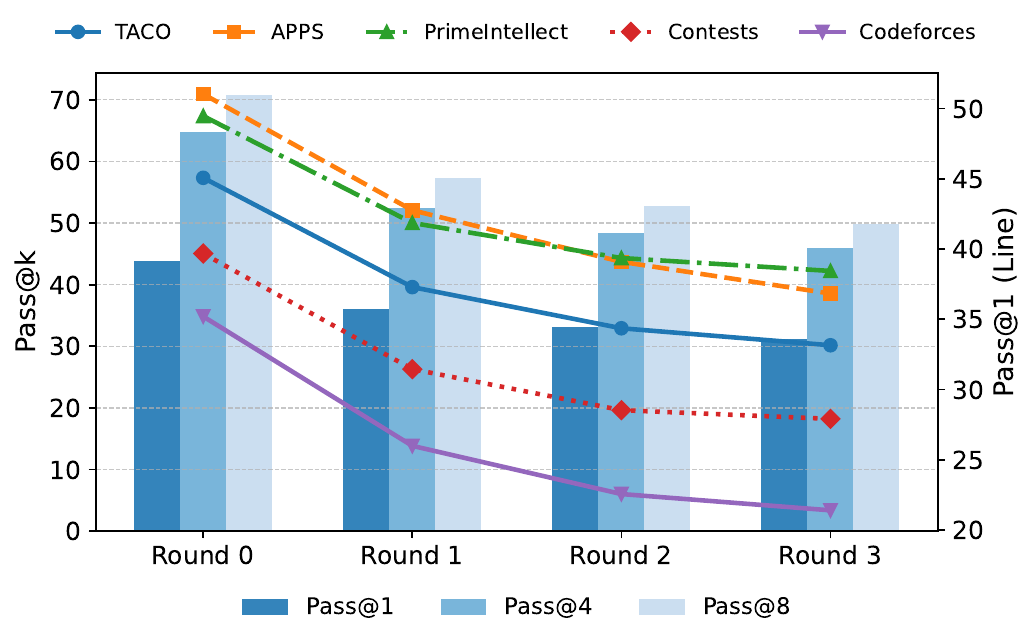}
  \caption{Difficulty evolution across rounds.}
  \label{fig:difficulty_evolution}
  \vspace{0em}
\end{figure}

\begin{figure}[t]
  \centering
  \includegraphics[width=\linewidth]{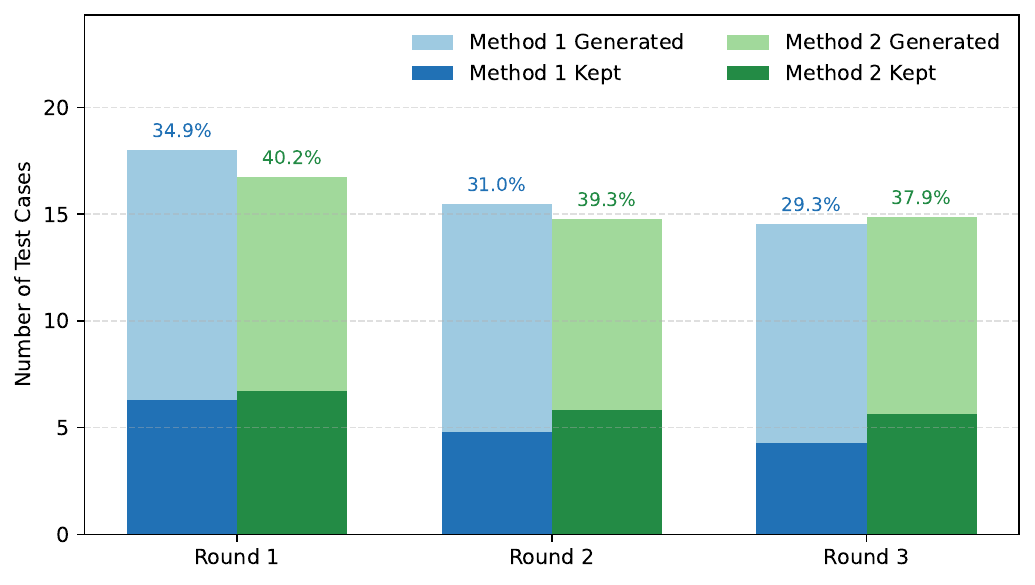}
  \caption{Generated and retained test cases across rounds for two generation strategies: Method 1 (Adversarial) and Method 2 (Discriminative).}
  \label{fig:two_prompts}
  \vspace{0em}
\end{figure}

\begin{figure}[t]
  \centering
  \includegraphics[width=\linewidth]{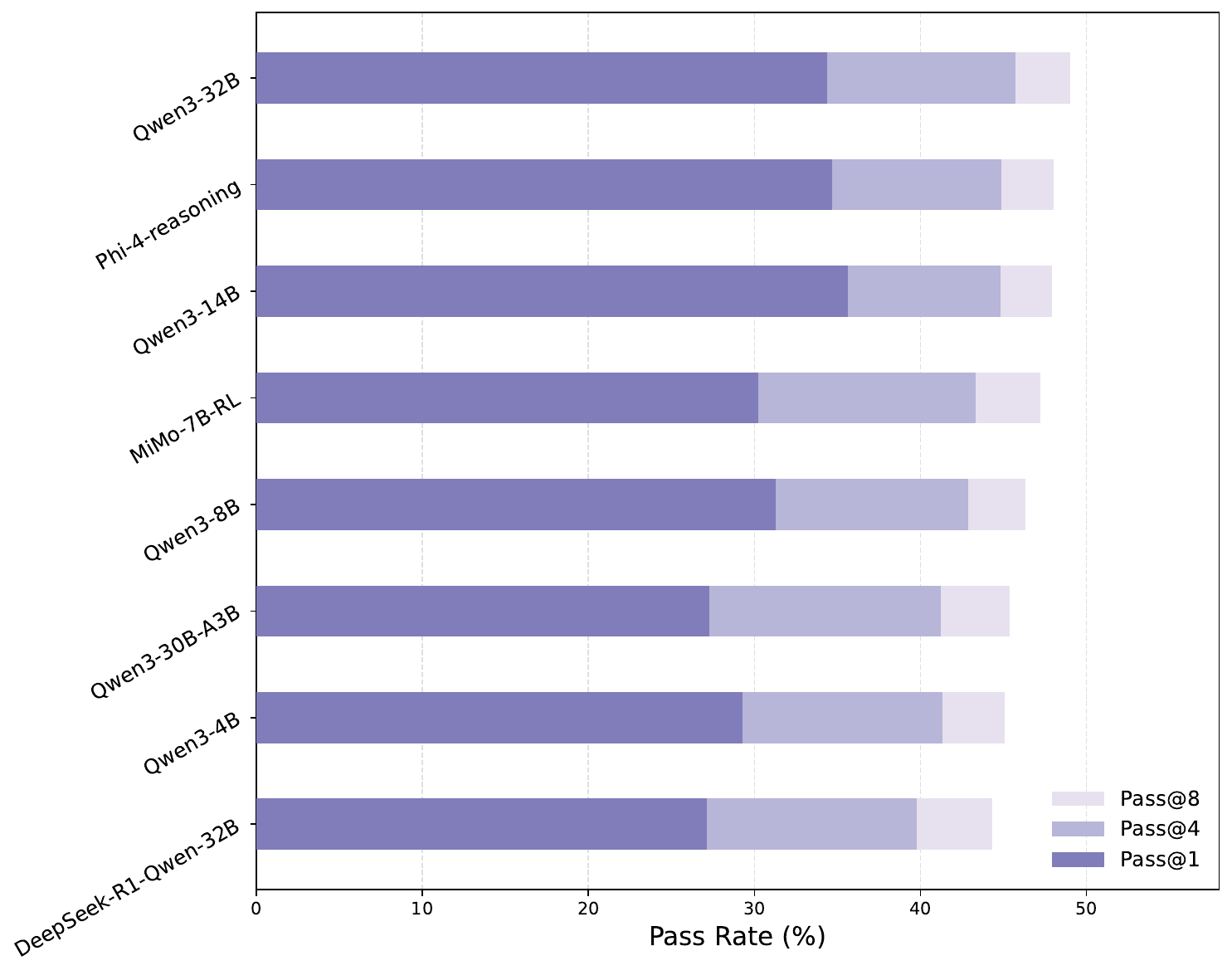}
  \caption{Pass@$k$ performance of diverse candidate solution models on the final dataset.}
  \label{fig:final_round_pass}
  \vspace{0em}
\end{figure}

\subsection{Difficulty Analysis}

We analyze how evaluation difficulty evolves across iterative construction rounds by measuring model Pass@$k$ under the test suite produced at each iteration, where evaluated solutions are uniformly sampled from the full candidate pool $\mathcal{S}(q) = \{ s_i \}_{i=1}^{64}$. While the underlying problems remain unchanged, later rounds strengthen verification by introducing increasingly adversarial and fine-grained corner test cases. As shown in Figure~\ref{fig:difficulty_evolution}, Pass@1, Pass@4, and Pass@8 decrease monotonically from Round~0 to Round~3, indicating that progressively refined test suites reject a larger fraction of partially correct solutions and expose failure modes that were previously undetected. Although a decrease in Pass@$k$ alone does not necessarily imply improved verification quality, in our setting it is accompanied by increased discrimination and stable RL gains.

Building on this analysis, we further examine Pass@$k$ performance of different candidate solution models on the final-round test suite, as shown in Figure~\ref{fig:final_round_pass}. Clear performance stratification emerges across models under the same, fully refined verification protocol. As expected, Pass@1, Pass@4, and Pass@8 increase with larger $k$ for all models. However, substantial gaps persist in absolute performance: larger-scale or reasoning-enhanced models, such as Qwen3-14B, Qwen3-32B, and Phi-4-Reasoning, achieve consistently higher pass rates across all $k$. This dominance suggests that increased parameter count and specialized reasoning architectural priors significantly bolster a model's ability to navigate the complex constraints of the final dataset. These models reflect stronger robustness to adversarial corner cases, whereas smaller and distilled models exhibit lower Pass@1 and Pass@4, revealing more frequent latent failures.

Notably, the performance differences across models narrow at higher $k$, particularly for Pass@8, suggesting that the final test suite is not merely overly restrictive but instead effectively differentiates single-solution reliability from multi-sample search capability. This behavior indicates that while weaker models can occasionally recover correct solutions through repeated sampling, stronger models are more likely to produce correct and robust solutions on the first attempt. Overall, these results demonstrate that the final dataset provides a balanced yet discriminative evaluation setting, enabling meaningful comparison of candidate solution models and serving as a reliable benchmark for downstream training and analysis.





\begin{table*}[t]
\centering
\footnotesize
\setlength{\tabcolsep}{0.5em}
\renewcommand{\arraystretch}{0.95}
\caption{Performance across rounds. Pass@1 results for models trained from Qwen3-4B (Thinking) using data from different refinement rounds, compared with strong reference models. We visualize gains of \coder{} (r3) to each baseline in the \colorbox{DeltaBg}{$\Delta$ column}. \textit{Abbreviations:} BCB=BigCodeBench-Instruct, LCB=LiveCodeBench.}
\resizebox{\textwidth}{!}{%
\begin{tabular}{@{}l|cccccc|c@{}}
\toprule
\textbf{Model} & 
  \textbf{EvalPlus} & 
  \textbf{BCB-Full} & 
  \textbf{BCB-Hard} & 
  \textbf{Aider-Polyglot} & 
  \textbf{LCB-v5} & 
  \textbf{AVG} &
  \textbf{$\Delta$} \\
\midrule
AceCoder-7B & 82.7 & 43.3 & 19.6 & - & - & - & - \\
DeepSeek-R1-Distill-14B & 82.4 & 38.1 & 20.9 & 18.6 & 53.0 & 42.6 & \mygreen{15}{\gain{15\%}} \\
DeepCoder-14B & 85.3 & 38.2 & 18.2 & 18.4 & 60.6 & 44.1 & \mygreen{11}{\gain{11\%}} \\
DeepSeek-V2.5-238B & 83.8 & 48.9 & 27.0 & 17.8 & 42.6 & 44.0 & \mygreen{11}{\gain{11\%}} \\
Qwen3-4B (Thinking) & 85.2 & 42.0 & 20.9 & 21.8 & 54.2 & 44.8 & \mygreen{9}{\gain{9\%}} \\
Critique-Coder-4B & 86.5 & 43.1 & 23.0 & 24.4 & 59.0 & 47.2 & \mygreen{4}{\gain{4\%}} \\
o1 & 88.6 & 50.4 & 28.4 & 61.7 & 59.5 & 57.7 & - \\
\midrule
\coder (r0) & 88.3 & 41.2 & 23.0 & 24.4 & 56.0 & 46.6 & \mygreen{5}{\gain{5\%}} \\
\coder (r1) & 88.8 & 42.4 & 23.0 & 24.4 & 57.2 & 47.2 & \mygreen{4}{\gain{4\%}} \\
\coder (r2) & 88.8 & 42.8 & 24.3 & 27.1 & 59.6 & 48.5 & \mygreen{1}{\gain{1\%}} \\
\rowcolor[HTML]{FFCCC9}
\textbf{\coder (r3)} & 
  \textbf{89.0} & 
  \textbf{43.4} & 
  \textbf{25.0} & 
  \textbf{27.4} & 
  \textbf{59.7} & 
  \textbf{49.0} &
  - \\
\bottomrule
\end{tabular}
}
\label{tab:round_eval}
\end{table*}

\subsection{Comparison of Test Case Generation Methods}


This section compares two generation methods based on their output volume and retention rates across iterative rounds. We use pass-vector diversity as a proxy for discriminative power. As shown in Figure~\ref{fig:two_prompts}, both methods show a downward trend in generation volume over iterations, reflecting a convergence toward more targeted test cases as verification refines.

Despite similar generation trends, Method 2 consistently achieves higher retention rates, indicating its tests are more effective at surviving filtering and discriminating between solutions. In contrast, the alternative method exhibits higher redundancy, with more tests removed during filtering. These results highlight that generation strategy significantly impacts test suite efficiency even under identical refinement protocols.






\section{Performance Evaluation}

In this section, we evaluate our dataset through controlled empirical studies. We compare models trained on different construction rounds to analyze how refined verification signals influence learning outcomes. We further benchmark our models against strong baselines to assess performance on the final dataset.

\subsection{Experimental Settings}

\textbf{Training Setup.} We adopt Group Relative Policy Optimization (GRPO) for reinforcement learning. Unless otherwise specified, we use a group size of 8 and apply nucleus sampling with temperature 0.6 and top-p 0.95 during rollout generation. All experiments are conducted using the Qwen3-4B model, with a maximum prompt length of 4,096 tokens and a maximum response length of 32,768 tokens. Policy optimization is performed with a learning rate of $1\times10^{-6}$. We set the upper and lower clipping ratios of the actor to 0.30 and 0.20, respectively, to stabilize updates. All models are trained under the same optimization settings to ensure fair comparison.

\textbf{Benchmarks.} We evaluate our trained models on four widely used coding benchmarks: EvalPlus~\citep{liu2023your}, which aggregates HumanEval~\citep{chen2021evaluating}, HumanEval+, MBPP~\citep{austin2021program}, and MBPP+, BigCodeBench-Instruct~\citep{zhuo2024bigcodebench}, Aider-Polyglot~\citep{aider_polyglot}, and LiveCodeBench v5 (2024.10--2025.02)~\citep{jain2024livecodebench}. These benchmarks cover a diverse range of programming tasks and difficulty levels, enabling a comprehensive assessment of code generation and reasoning performance across different evaluation regimes.

\textbf{Evaluation Setup.} For evaluation, we follow the thinking-mode sampling configuration reported in the original Qwen3 paper~\citep{yang2025qwen3}, using a temperature of 0.6, top-p 0.95, top-k 20, and a maximum generation length of 32,768 tokens, which is applied consistently across all benchmarks. For LiveCodeBench, we adopt the official thinking-mode evaluation prompt. We additionally compare our models against several strong coding baselines, including DeepSeek-R1-Distill-14B~\citep{guo2025deepseek}, DeepCoder~\citep{deepcoder2025}, DeepSeek-V2.5~\citep{deepseekv2}, and GPT-o1~\citep{o1systemcard2024}, evaluated under high-reasoning settings where applicable.

\subsection{Results}

Table \ref{tab:round_eval} summarizes model performance across iterative reinforcement learning rounds under the same training and evaluation settings. We observe steady and consistent improvements as training progresses from Round 0 to Round 3 across all benchmarks. In particular, gains are most pronounced on BigCodeBench-I, Aider-Polyglot, and LiveCodeBench, indicating that later rounds lead to stronger generalization on harder and more diverse coding tasks. By Round 3, the model achieves the best overall results, reaching an average score of 49.0, representing a +2.4 improvement over the Round 0 model.

Compared with strong baselines, the final model trained in Round 3 consistently outperforms Qwen3-4B (Thinking) and Critique-Coder-4B across all evaluation metrics, and remains competitive even compared with substantially larger models such as DeepSeek-V2.5-238B. These results demonstrate that iterative reinforcement learning alone, when applied over progressively refined training rounds, yields systematic and reliable performance gains without increasing model scale, improving both overall coding ability and robustness on challenging benchmarks.

\section{Related Works}

\subsection{Unit Tests Generation}
Unit testing plays an essential role in verifying the correctness of the generated solution. CodeT \citep{chen2022codet} and MPSC \citep{huang2023enhancing} made early attempts to leverage LLM to generate both solutions and test cases, selecting the best solution from multiple samples based on execution results on test cases. \citep{yang2024evaluation} systematically investigated the effectiveness of LLMs in generating unit tests through extensive empirical studies. AceCoder \citep{zeng2025acecoder} synthesized test cases along with questions from seed datasets in LeetCode style, then utilized Qwen2.5-Coder-32B-Instruct to perform quality control. KodCode \citep{xu2025kodcode} first used GPT-4o-0513 to generate both solutions and test cases, then employed a self-verification procedure to verify them.

\subsection{Synthetic Data Generation}
Since manually labeled data is costly to obtain, people have turned to LLM-generated data as an alternative solution. Self-instruct \citep{wang2022self} leveraged self-generated instruction data to enhance LLMs’ instruction-following capabilities. UltraChat \citep{ding2023enhancing} provided richly structured multi-turn instructional data, which plays a critical role in fostering general chat model capabilities. Dromedary \citep{sun2023principle} was developed by applying Self-align to LLaMA-65B, with alignment data playing a central role in its training. Evol-Instruct \citep{luo2023wizardcoder} utilized an evolutionary algorithm that generates diverse and complex instruction data for LLM.

\subsection{Reinforcement Learning for Coding Task}
Reinforcement learning has shown growing promise in the LLM post-training stage, attracting significant attention for its potential in code generation. CodeRL \citep{le2022coderl} introduced the first RL framework in code generation, utilizing unit test signals in an actor-critic architecture. After that, PPOCoder \citep{shojaee2023execution} extends this approach by integrating the PPO algorithm, and RLTF \citep{liu2023rltf} further refines it by providing feedback of multi-granularity to capture both syntactic and functional correctness. To enhance RL exploration in generating lengthy code, StepCoder \citep{dou2024stepcoder} adopts a step-by-step strategy, breaking down the task into manageable steps. Building upon these foundations, recent works have shifted focus toward more complex, real-world scenarios. SWE-RL \citep{wei2025swe} extends reinforcement learning to long-context, repository-level software engineering tasks.
\section{Conclusions}
\label{sec:conclusion}

Existing reinforcement learning with verifiable rewards (RLVR) for code generation is fundamentally limited by weak and static verification signals. In this work, we propose a solution-conditioned adversarial verification paradigm that refines test cases based on model execution behaviors, offering an alternative to fixed or single-pass test generation. This paradigm iteratively increases test difficulty and discriminative power while controlling redundancy and verification cost. Building on this approach, we introduce \dataset, a coding RL dataset with substantially stronger verification signals. Extensive experiments show that iterative adversarial refinement leads to more stable training and consistent performance gains across diverse coding benchmarks, highlighting the practical value of adversarial verification for effective code reinforcement learning.

\section*{Limitations}

\textbf{No Formal Guarantee on Problem and Test Correctness}

Despite heuristics such as filtering, cross-model validation, and execution-based consistency checks, our pipeline does not provide a formal guarantee of the correctness of generated problems and test cases. While methods are employed to remove invalid tests and problems, these procedures are empirical in nature. As a result, it remains possible that a small fraction of generated instances contain specification ambiguities, incomplete coverage, or incorrectness that are not exposed by the candidate solution pool. We therefore rely on empirical evidence, such as stable reinforcement learning dynamics and consistent performance improvements across benchmarks, to demonstrate the overall quality of the dataset, rather than offering theoretical guarantees. Establishing formal correctness or soundness guarantees for large-scale, automatically constructed coding datasets remains an open and challenging problem.

\noindent \textbf{Restriction to Python Programs}

Our dataset construction and evaluation are limited to Python programs. This choice is motivated by the availability of mature execution tooling, sandboxing infrastructure, and widely adopted benchmarks that enable reliable and large-scale execution-based verification. Python also provides a practical environment for validating reinforcement learning outcomes. While we expect the core principles of our pipeline to be largely language-agnostic, \dataset itself is not directly applicable to other programming languages. Extending the pipeline to additional languages would require language-specific execution environments, safety mechanisms, and benchmark support, which we leave to future work.


\bibliography{custom}

\appendix

\clearpage
\section{Appendix}
\label{sec:appendix}

\subsection{Prompt Template used for Creating Initial Round Dataset}
\label{appendix:seed_code_prompt}

\begin{promptbox}[title=Instruction for {LeetCode Question Generator}]{toolcardborder}
\textbf{Task:} Transform a code snippet into a very challenging LeetCode-style question intended for advanced CS university students and experienced software engineers.

\vspace{6pt}

\noindent\textbf{Role / Prompt:}
You are the latest and best bot aimed at transforming some code snippet into a very challenging LeetCode-style question intended for advanced CS university students and experienced software engineers.

\vspace{6pt}

\noindent\textbf{You will be given:}
\begin{enumerate}[left=0pt, itemsep=0pt, parsep=0pt]
    \item Original question/prompt for writing code
    \item Reference program that attempts to answer the question
\end{enumerate}

\vspace{6pt}

\noindent\textbf{Primary objectives (what to create):}
\begin{enumerate}[left=0pt, itemsep=0pt, parsep=0pt]
    \item Create a LeetCode-style question that meets the requirements below
    \item Generate 20 independent test cases using assert statements
\end{enumerate}

\vspace{6pt}

\noindent\textbf{Question requirements (must follow):}
\begin{enumerate}[left=0pt, itemsep=2pt, parsep=0pt]
    \item \textbf{Difficulty level:}
    \begin{itemize}[left=0pt, label={}, itemsep=0pt, parsep=0pt]
        \item The question must be hard or very hard difficulty level (similar to the hardest LeetCode problems).
        \item Challenging enough that solving it takes 30--60 minutes for experts.
    \end{itemize}

    \item \textbf{Algorithmic requirements:}
    \begin{itemize}[left=0pt, label={}, itemsep=0pt, parsep=0pt]
        \item The question should require advanced algorithmic thinking, such as:
        \begin{itemize}[left=12pt, label={$\rightarrow$}, itemsep=0pt, parsep=0pt]
            \item Graph theory with dynamic programming.
            \item Advanced string processing (suffix arrays, KMP, etc.).
            \item Complex greedy + data structure combinations.
            \item Sliding windows with optimization, interval DP, or segment trees.
        \end{itemize}
    \end{itemize}

    \item \textbf{Format requirements:}
    \begin{itemize}[left=0pt, label={}, itemsep=0pt, parsep=0pt]
        \item Must contain a function signature, rather than stdin/stdout style.
        \item Must be self-contained (no external resources or data).
        \item Avoid machine learning, OS-level concepts, or anything requiring system calls or file I/O.
        \item Do \textbf{NOT} request time/space complexity analysis or ask for test cases in the question text.
    \end{itemize}

    \item \textbf{Adaptation policy:}
    \begin{itemize}[left=0pt, label={}, itemsep=0pt, parsep=0pt]
        \item You can take inspiration from the reference code snippet, but you may discard parts of it if necessary to make the question cleaner and harder.
    \end{itemize}
\end{enumerate}

\vspace{6pt}

\noindent\textbf{Test case requirements:}
\begin{enumerate}[left=0pt, itemsep=0pt, parsep=0pt]
    \item Generate 20 independent test cases using assert statements.
    \item Each test case must:
    \begin{itemize}[left=12pt, label={$\rightarrow$}, itemsep=0pt, parsep=0pt]
        \item Use constant values (no randomness or external resource calls).
        \item Be independent of other test cases.
        \item Include both input parameters and expected output.
    \end{itemize}
\end{enumerate}

\vspace{6pt}

\noindent\textbf{Output format (JSON ONLY):}
Return a JSON object with the following keys ONLY.
\begin{itemize}[left=0pt, label={}, itemsep=0pt, parsep=0pt]
    \item Do not output any text outside JSON.
\end{itemize}

\begin{codebox}
{
  "question": "...",
  "tests": ["assert ...", "assert ..."]
}
\end{codebox}

\vspace{6pt}

\noindent\textbf{Inputs:}
\begin{itemize}[leftmargin=0pt, label={}, itemsep=0pt, parsep=0pt]
 \item{Original Question:} 
 \smalltt{\{instruction\}}
 \item{Reference Program:} 
 \begin{codebox}
```python
{program}
```
\end{codebox}
\end{itemize}
\end{promptbox}

\begin{promptbox}[title=Instruction for {LeetCode Question Generator}]{toolcardborder}
\textbf{Task:} Transform a code snippet into a very challenging LeetCode-style question intended for advanced CS university students and experienced software engineers.

\vspace{6pt}

\noindent\textbf{Role / Prompt:}
You are the latest and best bot aimed at transforming some code snippet into a very challenging LeetCode-style question intended for advanced CS university students and experienced software engineers. You will be provided with a prompt for writing code.

\vspace{6pt}

\noindent\textbf{Primary objectives:}
\begin{enumerate}[left=0pt, itemsep=0pt, parsep=0pt]
    \item Create a LeetCode-style question that meets these requirements:
    \begin{itemize}[left=0pt, label={}, itemsep=0pt, parsep=0pt]
        \item The question must be hard or very hard difficulty level (similar to the hardest LeetCode problems).
        \item The question should require advanced algorithmic thinking, such as:
        \begin{itemize}[left=0pt, itemsep=0pt, parsep=0pt]
            \item Graph theory with dynamic programming.
            \item Advanced string processing (suffix arrays, KMP, etc.).
            \item Complex greedy + data structure combinations.
            \item Sliding windows with optimization, interval DP, or segment trees.
        \end{itemize}
        \item The question must:
        \begin{itemize}[left=0pt, itemsep=0pt, parsep=0pt]
            \item Contain a function signature, rather than stdin/stdout style.
            \item Be self-contained (no external resources or data).
            \item Be challenging enough that solving it takes 30--60 minutes for experts.
            \item Avoid machine learning, OS-level concepts, or anything requiring system calls or file I/O.
        \end{itemize}
        \item Do \textbf{NOT} request time/space complexity analysis or ask for test cases in the question text.
        \item You can take inspiration from the reference code snippet, but you may discard parts of it if necessary to make the question cleaner and harder.
    \end{itemize}

    \item Based on the question you create:
    \begin{itemize}[left=0pt, label={}, itemsep=0pt, parsep=0pt]
        \item Generate 20 independent test cases using assert statements.
        \item Each test case must:
        \begin{itemize}[left=0pt, itemsep=0pt, parsep=0pt]
            \item Use constant values (no randomness or external resource calls).
            \item Be independent of other test cases.
            \item Include both input parameters and expected output.
        \end{itemize}
    \end{itemize}
\end{enumerate}

\vspace{6pt}

\noindent\textbf{Output format (JSON ONLY):}
Return a JSON object with the following keys ONLY.
\begin{itemize}[left=0pt, label={}, itemsep=0pt, parsep=0pt]
    \item \smalltt{question}: The LeetCode-style question text (string).
    \item \smalltt{tests}: Array of assert statements (list of strings).
    \item Do not output any text outside JSON.
\end{itemize}

\begin{codebox}
{
  "question": "...",
  "tests": ["assert ...", "assert ..."]
}
\end{codebox}

\vspace{6pt}

\noindent\textbf{Inputs:}
\begin{itemize}[leftmargin=0pt, label={}, itemsep=0pt, parsep=0pt]
 \item{Original Question:} 
 \smalltt{<<instruction>>}
\end{itemize}
\end{promptbox}

\subsection{Prompt Template used for Adversarial Test Case Generation}
\label{appendix:adversarial_prompt}

\begin{promptbox}[title=Instruction for {Adversarial Test Case Generator}]{toolcardborder}
\textbf{Task:} Generate adversarial and diverse test cases that reveal subtle weaknesses in high-performing programs. Create 20 new assert-based test cases that significantly upgrade the current test suite.

\vspace{6pt}

\noindent\textbf{Role / Prompt:}
You are an advanced AI system specialized in generating adversarial and diverse test cases that reveal subtle weaknesses in high-performing programs. You will receive a coding problem, five different programs solving it, existing test cases, and their evaluation results.

\vspace{6pt}

\noindent\textbf{You will be given:}
\begin{enumerate}[left=0pt, itemsep=0pt, parsep=0pt]
    \item Coding problem description (question)
    \item Five different programs solving the problem (Python code)
    \item Existing test cases
    \item Evaluation results (rows = programs, columns = tests)
\end{enumerate}

\vspace{6pt}

\noindent\textbf{Primary objectives (what to achieve):}
Create 20 new test cases that satisfy these requirements:
\begin{enumerate}[left=0pt, itemsep=0pt, parsep=0pt]
    \item Focus on challenging high-pass-rate programs
    \item Design cases likely to make at least one top-performing solution fail
    \item Include diverse input patterns that challenge different dimensions of the problem space
    \item Avoid repetition of existing test cases or trivial variations
    \item Ensure all test cases are correct according to the problem definition
    \item Make test cases independent of other test cases
    \item Use constant values (no randomness or external resource calls)
    \item Include both input parameters and expected output
\end{enumerate}

\vspace{6pt}

\noindent\textbf{Hard constraints (must follow):}
\begin{enumerate}[left=0pt, itemsep=2pt, parsep=0pt]
    \item \textbf{Correctness requirement:}
    \begin{itemize}[left=0pt, label={}, itemsep=0pt, parsep=0pt]
        \item All test cases must be correct according to the problem definition.
        \item Do \textbf{NOT} create test cases based on any specific program's behavior.
    \end{itemize}

    \item \textbf{Diversity requirement:}
    \begin{itemize}[left=0pt, label={}, itemsep=0pt, parsep=0pt]
        \item Test cases must cover diverse input patterns across different dimensions of the problem space.
        \item Avoid duplicating existing test cases or creating trivial variations.
    \end{itemize}

    \item \textbf{Adversarial targeting:}
    \begin{itemize}[left=0pt, label={}, itemsep=0pt, parsep=0pt]
        \item Prioritize challenging high-pass-rate programs.
        \item Design cases likely to expose weaknesses in at least one top-performing solution.
    \end{itemize}

    \item \textbf{Test independence:}
    \begin{itemize}[left=0pt, label={}, itemsep=0pt, parsep=0pt]
        \item Each test case must be independent of other test cases.
        \item Use constant values only (no randomness or external resource calls).
    \end{itemize}

    \item \textbf{Format requirement:}
    \begin{itemize}[left=0pt, label={}, itemsep=0pt, parsep=0pt]
        \item All test cases must be in assert-based format.
        \item Include both input parameters and expected output in each test case.
    \end{itemize}
\end{enumerate}

\vspace{6pt}

\noindent\textbf{Output format (JSON ONLY):}
Return a JSON object with the following structure:
\begin{itemize}[left=0pt, label={}, itemsep=0pt, parsep=0pt]
    \item The output must be a JSON object with a single key \smalltt{tests}.
    \item The value must be an array of strings, where each string is an assert statement.
    \item Do not output any text outside JSON.
\end{itemize}

\begin{codebox}
{
  "tests": [
    "assert ...",
    "assert ...",
    ...
  ]
}
\end{codebox}

\vspace{6pt}

\noindent\textbf{Inputs:}
\begin{itemize}[leftmargin=0pt, label={}, itemsep=0pt, parsep=0pt]
 \item{Question:} 
 
 \smalltt{\{question\}}
 \item{Programs (5 Python programs):} 
 \smalltt{\{program1\}}
 
 \smalltt{\{program2\}}
 
 \smalltt{\{program3\}}
 
 \smalltt{\{program4\}}
 
 \smalltt{\{program5\}}
 
 \item{Existing tests:} 
 
 \smalltt{\{tests\}}
 \item{Evaluation results (rows = programs, columns = tests):} 
 
 \smalltt{\{eval\_tests\}}
\end{itemize}
\end{promptbox}

\subsection{Prompt Template used for Discriminative Test Case Generation}
\label{appendix:discriminative_prompt}

\begin{promptbox}[title=Prompt for {Discriminative Test Case Generation}]{toolcardborder}
\textbf{Task:} Generate differentiating test cases that expose logical and behavioral differences between similar programs.

\vspace{6pt}

\noindent\textbf{Role / Prompt:}
You are an advanced AI system specialized in generating differentiating test cases that expose logical and behavioral differences between similar programs. You will receive a coding problem, five programs that currently produce mostly similar evaluation results, existing test cases, and their evaluation results.

\vspace{6pt}

\noindent\textbf{Your task:}
Create 20 new assert-based test cases that maximize discrimination power among these programs.

\vspace{6pt}

\noindent\textbf{Requirements:}
\begin{enumerate}[left=0pt, itemsep=2pt, parsep=0pt]
    \item \textbf{Differentiation requirement:}
    \begin{itemize}[left=0pt, label={}, itemsep=0pt, parsep=0pt]
        \item Each test case must clearly \textbf{differentiate among programs} that share similar evaluation results.
        \item At least one program must fail in each test case.
        \item At least one program must pass in each test case.
        \item Test cases should expose different failure modes across the programs.
    \end{itemize}

    \item \textbf{Correctness and independence:}
    \begin{itemize}[left=0pt, label={}, itemsep=0pt, parsep=0pt]
        \item All test cases must be correct according to the problem definition, not based on any specific program.
        \item Use constant values (no randomness or external resource calls).
        \item Each test case must be independent of other test cases.
        \item Include both input parameters and expected output.
    \end{itemize}
\end{enumerate}

\vspace{6pt}

\noindent\textbf{Output format (JSON ONLY):}
Return a JSON object with the following structure:
\begin{itemize}[left=0pt, label={}, itemsep=0pt, parsep=0pt]
    \item Do not output any text outside JSON.
\end{itemize}

\begin{codebox}
{
  "tests": [
    "assert ...",
    "assert ..."
  ]
}
\end{codebox}

\vspace{6pt}

\noindent\textbf{Inputs:}
\begin{itemize}[leftmargin=0pt, label={}, itemsep=0pt, parsep=0pt]
 \item{Question:} 
 
 \smalltt{\{question\}}
 \item{Programs (5 Python programs):} 
 \smalltt{\{program1\}}
 
 \smalltt{\{program2\}}
 
 \smalltt{\{program3\}}
 
 \smalltt{\{program4\}}
 
 \smalltt{\{program5\}}
 
 \item{Existing tests:} 
 
 \smalltt{\{tests\}}
 \item{Evaluation results (rows = programs, columns = tests):} 
 
 \smalltt{\{eval\_tests\}}
 
\end{itemize}
\end{promptbox}

\end{document}